\documentclass[journal]{IEEEtran}

\usepackage{cite}
\usepackage{amsmath,amssymb,amsfonts}
\usepackage{algorithmic}
\usepackage{graphicx}
\usepackage{textcomp}
\usepackage{xcolor}
\usepackage{hyperref}
\usepackage{url}
\usepackage{booktabs}

\hypersetup{
    colorlinks=true,
    linkcolor=blue,
    urlcolor=blue,
    citecolor=blue
}

\renewcommand{\footnoterule}{%
    \kern -3pt
    \hrule width \columnwidth height 0.4pt
    \kern 2pt
}

\def\BibTeX{{\rm B\kern-.05em{\sc i\kern-.025em b}\kern-.08em
    T\kern-.1667em\lower.7ex\hbox{E}\kern-.125emX}}

\begin{document}

\title{Evidence Before Accuracy: A MRI-PET Fusion Network for Alzheimer's Disease Classification with Causal Regional Validation}

\author{Saeid~Firouzi~Daghigh
        and~Saeed~Ayat
\thanks{The authors are with the Department of Computer Engineering
and Information Technology, Payame Noor University, Tehran, Iran
(e-mail: saeedmr881@gmail.com; dr.ayat@pnu.ac.ir).}}

\maketitle

\begin{abstract}
Deep learning models for Alzheimer's disease (AD) classification routinely report near-perfect discrimination, yet few are shown to rest on AD-relevant neurobiology rather than on dataset artifacts, subject-level leakage, or non-brain image content. We present a six-stream 2.5D fusion network combining T1 MRI and FDG PET across axial, coronal, and sagittal planes, trained on a strictly subject-disjoint ADNI consists of 554 paired subjects. The fusion model reaches AUC 0.962, accuracy 0.909, and F1 0.891, competitive with recent 3D CNN and multimodal transformer systems at substantially lower cost. We first quantify how much modality, plane and slice geometry matter. A validation-only search over slice centres and neighbour spacings moves AUC by 0.180 for MRI and 0.078 for PET, selecting narrow spacing for MRI ($\Delta=4$\,mm) and wide spacing for PET ($\Delta=16$\,mm), with the chosen coronal centres falling on the
hippocampal body and on the posterior cingulate/precuneus respectively. The contribution, however, is the evidence layer built around that number. Shortcut controls collapse the model to AUC 0.622 (silhouette), 0.608 (exterior), and 0.500 (blank), and a label-permutation null yields 0.456. Forward region-of-interest (ROI) ablation shows that masking medial temporal cortex in MRI and the posterior default-mode network (DMN) in PET produces the largest shift in the AD logit, while area-matched controls remain indistinguishable from that null. Reverse ROI ablation shows that the medial temporal lobe alone retains 89.2\% of above-chance discrimination in MRI and the posterior DMN alone retains 79.0\% in PET, versus 31.6\% or less for area-matched and enlarged controls. A quantitative comparison of four attribution methods shows occlusion sensitivity reaching $3.5$-$5.0\times$ enrichment inside a priori AD regions against $0.10$-$0.43\times$ in controls. Ablation and attribution independently establish a biologically correct double dissociation: hippocampal evidence is carried by MRI, posterior cingulate and precuneus evidence by PET. We argue that control-anchored causal validation of this kind should be a reporting standard, and that gradient-based saliency alone is insufficient evidence of biological validity. Source code, pre-trained models, and test data results are publicly
available at: \url{https://github.com/saeed5959/ad}
\end{abstract}


\IEEEpeerreviewmaketitle

\section{Introduction}

\IEEEPARstart{A}{lzheimer's} disease affects tens of millions of people worldwide and remains
without a disease-modifying cure, which places a high value on accurate and
early characterisation of the disease state \cite{belhajali2025}. AD is diagnosed in vivo through a convergence of structural, metabolic, and molecular evidence. Structural MRI captures the medial temporal atrophy that tracks neurodegeneration and correlates with Braak staging \cite{frisoni2010}, while FDG-PET captures the temporoparietal and posterior cingulate hypometabolism that often precedes measurable volume loss \cite{nordberg2010}. Because the two modalities index different and only partially overlapping pathological processes, their combination improves detection and differential diagnosis relative to either alone \cite{dukart2011}. This complementarity is the standard motivation for multimodal deep learning in AD, and recent systematic reviews confirm that multimodal models consistently outperform unimodal ones across datasets and architectures \cite{yu2026}.

The methodological quality of that literature, however, is uneven. Reported accuracies above 0.98 are common, yet a substantial fraction are produced under conditions that make them uninterpretable: slice-level rather than subject-level splitting, which places slices from the same brain in both training and test sets; checkpoint selection on the test set; test sets of a few dozen subjects, for which the standard error of AUC alone is on the order of $\pm 0.04$; and evaluation on a single split, which makes cross-experiment comparison meaningless. Under these conditions a high number is evidence of a leak, not of learning.

A second and deeper problem is that even a correctly validated classifier may reach the right answer for the wrong reason. Medical imaging models are known to exploit acquisition signatures, head shape, skull thickness, field-of-view differences, and other confounds that correlate with class membership in a given cohort but carry no clinical meaning. The standard response has been to present saliency maps \cite{selvaraju2020,chattopadhyay2018,jiang2021}, but saliency is descriptive rather than causal: a heatmap indicates where gradients are large, not whether the model's decision depends on the tissue in that location. The explainability literature has repeatedly made this distinction between an explanation being produced and \emph{causability}, the degree to which an explanation supports genuine causal understanding by a clinician \cite{holzinger2019}, and has documented the gap between the volume of XAI methods proposed and the volume validated \cite{vandervelden2022,adadi2018,arrieta2020}. AD-specific XAI reviews reach the same conclusion: attribution maps are routinely produced and rarely tested \cite{khosroshahi2025}.

This work is organized around that gap. We build a deliberately conventional model, a six-stream VGG16-BN fusion network over tri-planar MRI and PET slices, and then subject it to an evidence ladder designed to answer four questions that a saliency map cannot:

\begin{enumerate}
\item \textbf{Does the model exploit non-brain shortcuts?} Answered by silhouette, exterior, and blank input controls, and by a label-permutation null.
\item \textbf{Does each modality contribute independently?} Answered by modality ablation on the fusion model and by preprocessing ablation.
\item \textbf{Are AD-relevant regions causally necessary and sufficient?} Answered by forward ROI ablation (mask the region) and reverse ROI ablation (keep only the region), both benchmarked against area-matched control regions and a spatial permutation null.
\item \textbf{Which attribution method actually localizes AD pathology?} Answered by quantitative enrichment and pointing-game analysis across occlusion, Layer-CAM, Grad-CAM, and Grad-CAM++.
\end{enumerate}

The classification result is therefore presented as a prerequisite rather than as the contribution. The contribution is the demonstration that a model of this class can be shown, with controls, to depend on hippocampal structure and posterior default-mode metabolism in the modality-specific pattern that AD biology predicts.

\section{Related Work}

\subsection{Multimodal MRI-PET Deep Learning}

Architectures for MRI-PET fusion divide broadly into early fusion at the image level \cite{song2021}, intermediate feature-level fusion, and late fusion at the decision level. Narazani \emph{et al.} \cite{narazani2022} systematically compared these strategies with 3D ResNets and reported that late fusion outperformed early and middle fusion, while raising the pointed question of whether PET alone is sufficient. Transformer-based approaches have followed: MMTFN introduces multi-scale transformer fusion across modalities \cite{miao2024}, DiaMond uses multi-modal vision transformers with explicit inter-modal attention \cite{li2025}, and MINiT applies multiple-instance transformers to neuroimaging volumes \cite{singla2022}. Vision transformers in general \cite{dosovitskiy2021} have been adapted to AD with mixed returns relative to their data requirements. Self-supervised and longitudinal formulations have also gained traction, including cross-modal MRI-PET pretraining with explicit site-invariance objectives \cite{belhajali2025}, contrastive pretraining on 3D amyloid PET for progression prediction \cite{kwak2023}, self-organized multi-modal longitudinal maps \cite{ouyang2024}, and multi-modal fusion with longitudinal analysis \cite{muksimova2025}. CNN-based work continues in parallel, including tri-branch EfficientNet integration \cite{prasun2026}, ensemble architectures \cite{desai2026}, and attention-driven CNNs with multi-activation fusion \cite{alsubaie2025}. Narrative reviews of the PET-MRI AI literature since Tauvid approval \cite{christodoulou2025} and systematic reviews of datasets and modalities \cite{yu2026} both note that architectural novelty has outpaced validation rigor.

\subsection{Explainability in AD Neuroimaging}

The AD-specific XAI literature spans post-hoc attribution, intrinsically interpretable models, and feature attribution on tabular biomarkers. El-Sappagh \emph{et al.} \cite{elsappagh2021} built a multilayer multimodal model with an explainability layer over clinical and imaging features; Coluzzi \emph{et al.} \cite{coluzzi2025} used XAI over multiple MRI-derived brain measures for biomarker investigation; Bhattarai \emph{et al.} \cite{bhattarai2024} applied Deep-SHAP to map regional neuroimaging biomarkers onto cognition; Hernandez \emph{et al.} \cite{hernandez2022} used XAI to interrogate the behavior of the top TADPOLE challenge methods. Anzum \emph{et al.} \cite{anzum2025} combined transformers with explainability, and Adeniran \emph{et al.} \cite{adeniran2025} proposed an explainable MRI-based ensemble architecture. Reviews \cite{khosroshahi2025,vandervelden2022} catalogue the methods; \cite{holzinger2019,adadi2018,arrieta2020} supply the conceptual framing, in particular the distinction between an explanation being produced and an explanation being \emph{verified}.

\subsection{Attribution Methods}

Occlusion sensitivity \cite{zeiler2014} perturbs input regions and measures the change in output, making it causal by construction but coarse and computationally expensive. Grad-CAM \cite{selvaraju2020} and Grad-CAM++ \cite{chattopadhyay2018} weight final-layer feature maps by gradients, producing smooth but low-resolution maps whose faithfulness depends strongly on the backbone and the layer chosen. Layer-CAM \cite{jiang2021} addresses the resolution limitation by combining hierarchical class activation maps from earlier layers. We evaluate all four under identical conditions rather than selecting the method that produces the most persuasive picture.

\subsection{Positioning}

Relative to this literature, the present work does not claim architectural novelty. It claims that the validation and explainability protocol reported here, shortcut controls, permutation nulls, forward and reverse regional ablation with area-matched controls, and quantitative attribution benchmarking, is more informative than an incremental AUC improvement, and that it exposes failure modes (notably the unreliability of Grad-CAM family methods in this setting) that are invisible to qualitative inspection.

\section{Materials and Methods}

\subsection{Dataset}

Data were obtained from the Alzheimer's Disease Neuroimaging Initiative (ADNI) \cite{petersen2010}. We used preprocessed T1-weighted structural MRI and FDG PET accepting all ADNI series to maximize the paired cohort.

The full download comprised 18{,}114 MRI series and 6{,}772 PET series. Restricting to subjects with both modalities available yielded 554 unique subjects: 324 cognitively normal (CN) and 230 AD. Splits were made \emph{at the subject level before any slice extraction}, producing 356 training, 88 validation, and 110 test subjects (Table~\ref{tab:cohort}). The test partition was physically separated into a distinct directory before training began and was not read by any script during model development, slice selection, or hyperparameter tuning.

\begin{table}[!t]
\renewcommand{\arraystretch}{1.2}
\caption{ADNI Cohort Composition}
\label{tab:cohort}
\centering
\begin{tabular}{lccc}
\toprule
Group & MRI -- PET series & Paired subjects & Train -- Val -- Test \\
\midrule
CN  & 12{,}198 -- 4{,}348 & 324 & 208 -- 52 -- 64 \\
AD  & 5{,}916 -- 2{,}424  & 230 & 148 -- 36 -- 46 \\
All & 18{,}114 -- 6{,}772 & 554 & 356 -- 88 -- 110 \\
\bottomrule
\end{tabular}
\end{table}

Subject-level splitting is not a refinement but a precondition. Splitting at the slice level permits slices from the same brain to appear on both sides of the partition, and in a 2.5D setting this reliably inflates reported metrics to the point where they measure memorization of individual anatomy rather than group discrimination.

\begin{figure}[!t]
\centering
\includegraphics[width=0.82\columnwidth]{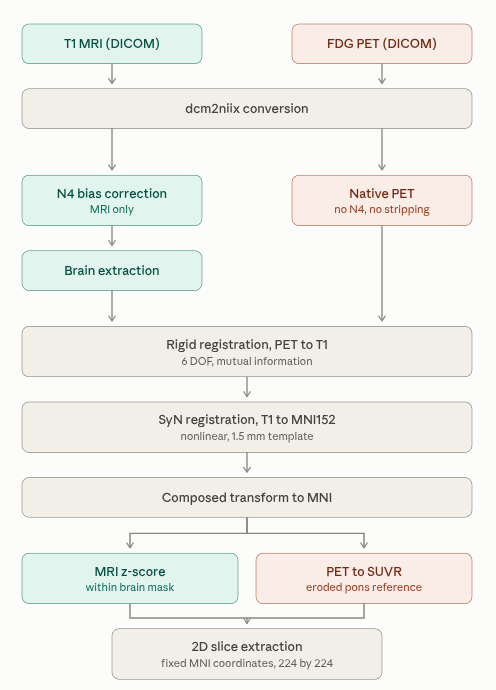}
\caption{Preprocessing pipeline. MRI and PET are converted from DICOM with
\texttt{dcm2niix}. MRI receives N4 bias correction and
brain extraction; PET is kept native, with no bias correction and
no skull stripping. PET is rigidly registered to the subject's T1 (6 DOF, mutual
information), the T1 is nonlinearly registered to the MNI152 1.5\,mm
template with ANTs SyN, and the two transforms are
composed so each volume is resampled exactly once. MRI is z-scored within the
brain mask; PET is converted to SUVR against an eroded pons
reference. Fixed MNI coordinates then define the extracted
$224\times224$ slices.}
\label{fig:pipeline}
\end{figure}

\subsection{Preprocessing}

\begin{figure*}[!t]
\centering
\includegraphics[width=0.86\textwidth]{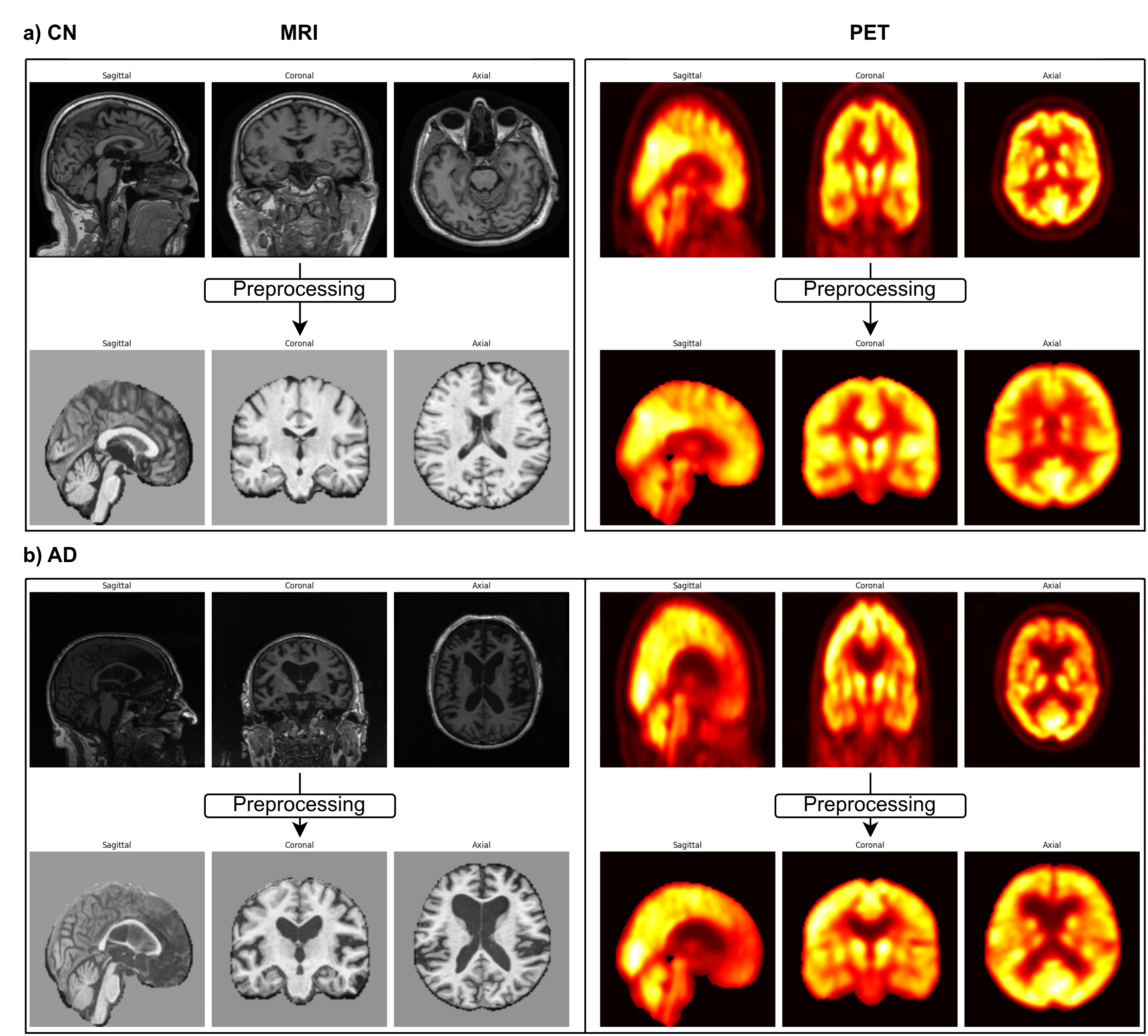}
\caption{Representative CN (a) and AD (b) subjects before and after
preprocessing, shown in all three planes for both modalities. The AD subject
shows the expected ventricular enlargement and cortical thinning on MRI, and
posterior hypometabolism on PET.}
\label{fig:examples}
\end{figure*}

The preprocessing pipeline is shown in Fig.~\ref{fig:pipeline}, with representative CN and AD examples before and after processing in Fig.~\ref{fig:examples}.

DICOM series were converted to NIfTI with \texttt{dcm2niix}. MRI volumes underwent N4 bias field correction followed by brain extraction with HD-BET \cite{isensee2019}. PET volumes were deliberately left native: no bias correction and no skull stripping, since PET has no comparable bias field and since applying an MRI-derived mask to PET risks removing genuine tracer signal at the cortical rim.

PET was rigidly registered to the subject's own T1 (6 degrees of freedom, mutual information), and the T1 was nonlinearly registered to the MNI152 1.5\,mm template with ANTs SyN. The two transforms were then \emph{composed} and applied in a single resampling step, so that each voxel is interpolated exactly once; sequential resampling would introduce avoidable blur that disproportionately affects the thin structures of interest.

Intensity normalization was modality-appropriate. MRI was $z$-scored within the brain mask. PET was converted to SUVR using an \emph{eroded pons} reference region drawn from the Harvard-Oxford subcortical atlas; erosion reduces partial-volume contamination at the reference boundary. Per-image min-max rescaling was explicitly avoided: for PET it destroys the inter-subject comparability that SUVR normalization exists to preserve, erasing precisely the global hypometabolism signal that distinguishes AD from CN.

Finally, 2D slices were extracted at fixed MNI coordinates and resampled to $224 \times 224$, stored as per-subject float32 arrays. Values were clipped to $(-3.0, 3.0)$ for both modalities and standardized with $\mu = 0.449$, $\sigma = 0.226$ to match ImageNet backbone statistics.

\subsubsection{Does Nonlinear Registration Remove the Atrophy Signal?}
This is the standard objection to SyN normalization in AD work. We tested it directly: CSF fraction computed in native space and in MNI space correlated at $r = 0.976$ across all 554 subjects, between-subject spread was retained at 0.902, and the mean absolute log-Jacobian of the warp was near-identical between groups (AD 0.198, CN 0.192). Morphological signal survives the warp; the transform is not selectively compressing AD brains toward the template.

\subsection{Slice Selection}

Because each stream consumes three adjacent slices stacked as a $(3, 224, 224)$ tensor, two parameters define a stream: the center coordinate $c$ in MNI millimeters and the inter-slice spacing $\Delta$. These were selected by grid search \emph{on the validation set only}, with the test set untouched (Tables~\ref{tab:mri_slices} and \ref{tab:pet_slices}).

\begin{table}[!t]
\renewcommand{\arraystretch}{1.1}
\caption{MRI Slice Search, Validation AUC}
\label{tab:mri_slices}
\centering
\small
\begin{tabular}{ccc@{\hskip 8pt}ccc@{\hskip 8pt}ccc}
\toprule
\multicolumn{3}{c}{Axial} & \multicolumn{3}{c}{Coronal} & \multicolumn{3}{c}{Sagittal} \\
\cmidrule(lr){1-3}\cmidrule(lr){4-6}\cmidrule(lr){7-9}
$c$ & $\Delta$ & AUC & $c$ & $\Delta$ & AUC & $c$ & $\Delta$ & AUC \\
\midrule
$+20$ & 0  & 0.704 & $-10$ & 0  & \textbf{0.884} & $-10$ & 0  & 0.758 \\
$+20$ & 4  & 0.782 & $-10$ & 4  & 0.871 & $-10$ & 4  & 0.781 \\
$+20$ & 8  & 0.753 & $-10$ & 8  & 0.814 & $-10$ & 8  & 0.826 \\
$+20$ & 14 & 0.742 & $-10$ & 14 & 0.848 & $-10$ & 14 & 0.798 \\
$0$   & 0  & 0.741 & $-25$ & 0  & 0.809 & $-25$ & 0  & 0.848 \\
$0$   & 4  & 0.786 & $-25$ & 4  & 0.816 & $-25$ & 4  & \textbf{0.860} \\
$0$   & 8  & 0.793 & $-25$ & 8  & 0.808 & $-25$ & 8  & 0.814 \\
$0$   & 14 & 0.751 & $-25$ & 14 & 0.778 & $-25$ & 14 & 0.764 \\
$-20$ & 0  & 0.842 & $-40$ & 0  & 0.768 & $-40$ & 0  & 0.769 \\
$-20$ & 4  & \textbf{0.875} & $-40$ & 4  & 0.772 & $-40$ & 4  & 0.783 \\
$-20$ & 8  & 0.812 & $-40$ & 8  & 0.782 & $-40$ & 8  & 0.802 \\
$-20$ & 14 & 0.816 & $-40$ & 14 & 0.765 & $-40$ & 14 & 0.784 \\
\bottomrule
\end{tabular}
\end{table}

\begin{table}[!t]
\renewcommand{\arraystretch}{1.1}
\caption{PET Slice Search, Validation AUC}
\label{tab:pet_slices}
\centering
\small
\begin{tabular}{ccc@{\hskip 8pt}ccc@{\hskip 8pt}ccc}
\toprule
\multicolumn{3}{c}{Axial} & \multicolumn{3}{c}{Coronal} & \multicolumn{3}{c}{Sagittal} \\
\cmidrule(lr){1-3}\cmidrule(lr){4-6}\cmidrule(lr){7-9}
$c$ & $\Delta$ & AUC & $c$ & $\Delta$ & AUC & $c$ & $\Delta$ & AUC \\
\midrule
$+20$ & 0  & 0.875 & $-40$ & 0  & 0.923 & $-6$  & 0  & 0.925 \\
$+20$ & 6  & 0.908 & $-40$ & 6  & 0.926 & $-6$  & 6  & 0.928 \\
$+20$ & 10 & 0.901 & $-40$ & 10 & \textbf{0.947} & $-6$ & 10 & 0.928 \\
$+20$ & 16 & 0.905 & $-40$ & 16 & 0.943 & $-6$  & 16 & 0.931 \\
$+32$ & 0  & 0.916 & $-55$ & 0  & 0.893 & $-25$ & 0  & 0.901 \\
$+32$ & 6  & 0.904 & $-55$ & 6  & 0.911 & $-25$ & 6  & 0.923 \\
$+32$ & 10 & 0.909 & $-55$ & 10 & 0.908 & $-25$ & 10 & 0.906 \\
$+32$ & 16 & 0.911 & $-55$ & 16 & 0.928 & $-25$ & 16 & \textbf{0.940} \\
$+42$ & 0  & 0.869 & $-70$ & 0  & 0.893 & $-45$ & 0  & 0.886 \\
$+42$ & 6  & 0.902 & $-70$ & 6  & 0.896 & $-45$ & 6  & 0.881 \\
$+42$ & 10 & 0.917 & $-70$ & 10 & 0.911 & $-45$ & 10 & 0.910 \\
$+42$ & 16 & \textbf{0.925} & $-70$ & 16 & 0.904 & $-45$ & 16 & 0.914 \\
\bottomrule
\end{tabular}
\end{table}

The selected configurations are anatomically coherent rather than arbitrary. The MRI streams converge on the medial temporal lobe: axial $c = -20$\,mm passes through the hippocampal body and amygdala, coronal $c = -10$\,mm through the hippocampal head, and sagittal $c = -25$\,mm along the long axis of the hippocampus. Narrow spacing ($\Delta = 4$\,mm) preserves the fine structural detail on which atrophy assessment depends. The PET streams converge on the posterior default-mode network: coronal $c = -40$\,mm and sagittal $c = -25$\,mm intersect the posterior cingulate and precuneus, and axial $c = +42$\,mm covers the parietal association cortex. Wide spacing ($\Delta = 16$\,mm) is appropriate for PET because metabolic deficits are spatially diffuse and the effective resolution is far coarser than that of T1.

Notably, the MRI grid shows strong sensitivity to slice position (validation AUC ranges from 0.704 to 0.884), whereas the PET grid is comparatively flat (0.869 to 0.947). This is itself informative: metabolic AD signal is distributed widely enough that most posterior slices carry it, while structural signal is concentrated in a narrow band around the hippocampus.

\subsection{Architecture}

\begin{figure}[!t]
\centering
\includegraphics[width=\columnwidth]{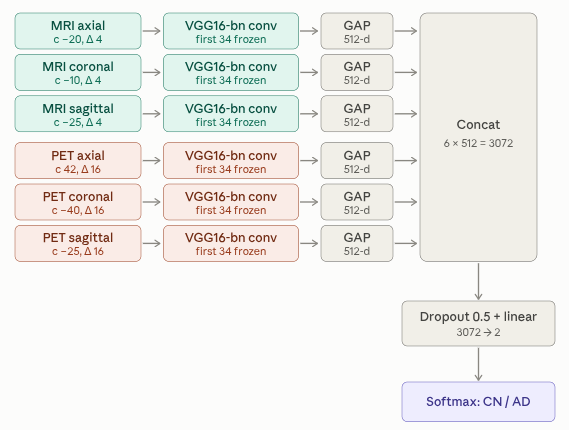}
\caption{Six-stream fusion architecture. Each of the six inputs (MRI and PET
$\times$ three planes) passes through an independent VGG16-BN
convolutional trunk initialised from ImageNet, with the first 34
layers frozen. Global average pooling produces a 512-dimensional
descriptor per stream. The six descriptors are concatenated into a
3072-dimensional vector, followed by dropout ($p=0.5$) and a single linear layer
to two logits.}
\label{fig:arch}
\end{figure}

The fusion network shown in Fig.~\ref{fig:arch} consists of six parallel streams, MRI axial, MRI coronal, MRI sagittal, PET axial, PET coronal, PET sagittal, each processing its own $(3, 224, 224)$ input.

Each stream uses a VGG16-BN convolutional trunk with the \emph{first 34 layers frozen} and later layers fine-tuned. Freezing early layers preserves generic edge and texture filters while limiting the number of trainable parameters relative to a cohort of 554 subjects; it is a direct response to the sample size rather than an architectural preference. Each trunk is followed by global average pooling to a 512-dimensional descriptor, replacing VGG's fully connected head and eliminating the large majority of its parameters.

The six 512-dimensional descriptors are concatenated into a 3072-dimensional joint representation, passed through dropout ($p = 0.5$), and mapped by a single linear layer to two logits with softmax over $\{$CN, AD$\}$.

\subsubsection{Stream Initialization}
Models are built in three stages, and the initialization of each stage is what distinguishes the conditions reported in Tables~\ref{tab:mri_test}-~\ref{tab:pet_test}-\ref{tab:fusion_test}. Single-plane models are initialized from ImageNet weights and trained on one plane of one modality. Tri-planar models are then trained in two variants: in the first condition all three trunks are again initialized from ImageNet, whereas in the \emph{single-plane init} condition each trunk is initialized from the corresponding trained single-plane model of the same modality and orientation, and the three-stream network is subsequently trained end-to-end. Finally, the six-stream multimodal model inherits its trunks from the two trained tri-planar models and is trained end-to-end in the same way. The layer-freezing policy is identical across all stages.

Transferring single-plane weights rather than ImageNet weights into the tri-planar model gives each stream a plane-specific starting point already adapted to MNI-space neuroimaging intensities, and is the single largest contributor to the MRI and PET result.

\subsection{Training}

Models were implemented in PyTorch and trained with cross-entropy loss. Checkpoints were selected on \emph{validation AUC}, never on test performance. All reported figures derive from complete 5-fold cross-validation.
 
Optimization used AdamW with a learning rate of $1\times10^{-5}$, a batch size of 16, and 60 epochs. The sole augmentation was a random affine transform (rotation up to $5^{\circ}$, translation up to 3\% of image extent in each axis), applied to the training split only.

\subsection{Explainability Protocol}

Four analyses were applied to the trained fusion model on the held-out test set.

\subsubsection{Shortcut Controls}
Four input conditions were compared: \emph{intact}; \emph{silhouette}, in which brain tissue is replaced by a uniform fill so that only the outer contour remains; \emph{exterior}, in which the brain is removed and only non-brain content is retained; and \emph{blank}. If a model is reading head shape, skull thickness, or field-of-view artifacts, silhouette and exterior conditions will retain discriminative power.

\subsubsection{Label-Permutation Null}
The full training pipeline was rerun with class labels shuffled within the training set, preserving all other properties of the data. This estimates the discrimination achievable by the architecture and training procedure in the absence of real class structure, and thereby bounds the contribution of optimization artifacts.

\subsubsection{Forward and Reverse ROI Ablation}
Regions of interest were taken from the Harvard-Oxford cortical and subcortical atlases and projected into the $224 \times 224$ MNI slice space of each stream. \emph{Forward ablation} masks the ROI and measures the resulting change in AUC and in the AD logit. \emph{Reverse ablation} inverts the operation: everything outside the ROI is masked and the model is evaluated on the ROI alone, measuring sufficiency rather than necessity. Both are compared against area-matched control composites and against a spatial permutation null in which masks of identical area are placed at random brain locations; the resulting $Z$ statistic expresses effect size relative to that null.

Reverse ablation was necessary because the six-stream architecture is highly redundant. Forward ablation of a single small region in a single stream produces near-zero $\Delta$AUC simply because five other streams remain intact, a property of the model, not evidence of regional irrelevance. Necessity tests alone systematically understate regional importance in redundant architectures.

\subsubsection{Attribution Benchmarking}
Occlusion sensitivity \cite{zeiler2014}, Layer-CAM \cite{jiang2021}, Grad-CAM \cite{selvaraju2020}, and Grad-CAM++ \cite{chattopadhyay2018} were computed per stream. Each was evaluated quantitatively by (i) \emph{enrichment}, the ratio of mean attribution inside an a priori ROI to the mean attribution expected under chance, computed separately for AD regions and area-matched control regions; and (ii) a \emph{pointing game}, the fraction of subjects for which the single peak attribution voxel falls inside the a priori AD ROI set, compared against the chance rate implied by that ROI set's area fraction. Both statistics were computed separately for AD and CN subjects, since a method that highlights the same regions regardless of class is not explaining the decision.

\section{Results}

\subsection{Single-Plane, Tri-Planar, and Multimodal Performance}

\begin{table}[!t]
\renewcommand{\arraystretch}{1.2}
\caption{MRI Performance, Test Set}
\label{tab:mri_test}
\centering
\begin{tabular}{lccc}
\toprule
Mode & AUC & Accuracy & F1 \\
\midrule
Axial    & 0.877 & 0.736 & 0.695 \\
Coronal  & 0.891 & 0.764 & 0.735 \\
Sagittal & 0.897 & 0.845 & 0.813 \\
Fuse 3-plane & 0.909 & 0.818 & 0.778 \\
Fuse 3-plane single-plane init & \textbf{0.944} & \textbf{0.873} & \textbf{0.844} \\
\bottomrule
\end{tabular}
\end{table}

\begin{table}[!t]
\renewcommand{\arraystretch}{1.2}
\caption{PET Performance, Test Set}
\label{tab:pet_test}
\centering
\begin{tabular}{lccc}
\toprule
Mode & AUC & Accuracy & F1 \\
\midrule
Axial    & 0.913 & 0.836 & 0.809 \\
Coronal  & 0.900 & 0.818 & 0.800 \\
Sagittal & 0.918 & 0.827 & 0.796 \\
Fuse 3-plane & 0.928 & 0.827 & 0.796 \\
Fuse 3-plane single-plane init & \textbf{0.936} & \textbf{0.855} & \textbf{0.830} \\
\bottomrule
\end{tabular}
\end{table}

\begin{table}[!t]
\renewcommand{\arraystretch}{1.2}
\caption{Multimodal Fusion, Test Set}
\label{tab:fusion_test}
\centering
\begin{tabular}{lccc}
\toprule
Mode & AUC & Accuracy & F1 \\
\midrule
MRI 3-plane & 0.944 & 0.873 & 0.844 \\
PET 3-plane & 0.936 & 0.855 & 0.830 \\
Fusion MRI + PET & \textbf{0.962} & \textbf{0.909} & \textbf{0.891} \\
\bottomrule
\end{tabular}
\end{table}

Three observations follow from Tables~\ref{tab:mri_test}-\ref{tab:pet_test}-\ref{tab:fusion_test}. First, PET outperforms MRI at the single-plane level in every plane (0.900--0.918 versus 0.877--0.897), consistent with the finding that metabolic signal is the stronger single discriminator in AD versus CN classification \cite{narazani2022}.

Second, the planes are not equally informative in the two modalities. For MRI, the sagittal plane is clearly the strongest (accuracy 0.845), ahead of coronal (0.764) and axial (0.736). For PET, by contrast, the three planes perform comparably (accuracy 0.836, 0.818, and 0.827 for axial, coronal, and sagittal respectively), with no plane carrying a decisive advantage. This mirrors the pattern already seen in the slice search (Tables~\ref{tab:mri_slices} and \ref{tab:pet_slices}): structural AD signal is concentrated along the long axis of the hippocampus and is therefore strongly plane-dependent, whereas metabolic AD signal is spatially diffuse and is captured similarly well from any of the three orientations.

Third, initialization matters for the tri-planar models. In the ``no pretrain'' condition the three streams are initialized from ImageNet weights; in the ``single-plane init'' condition each stream is instead initialized from the weights of the corresponding single-plane model, and the fusion network is then trained end-to-end. This second scheme improves both modalities, substantially so for MRI ($0.909 \rightarrow 0.944$ AUC, $0.818 \rightarrow 0.873$ accuracy) and more modestly for PET ($0.928 \rightarrow 0.936$). In both cases the pretrained tri-planar model also exceeds the best single plane, confirming that the three views contribute complementary rather than redundant information. Multimodal fusion adds a further gain on top of that: combining the MRI and PET tri-planar models yields AUC 0.962, accuracy 0.909, and F1 0.891, improving on the better unimodal model by 0.018 AUC, 0.036 accuracy, and 0.047 F1. That the gain is larger in the threshold-dependent metrics than in AUC indicates that fusion principally corrects subjects near the decision boundary rather than reordering the ranking wholesale.

\subsection{Comparison with Published Methods}

\begin{table}[!t]
\renewcommand{\arraystretch}{1.2}
\caption{Comparison with Published Models}
\label{tab:sota}
\centering
\begin{tabular}{lccc}
\toprule
Method & Modality & AUC & BACC \\
\midrule
3D-ResNet \cite{narazani2022} & M & 0.936 & 0.865 \\
3D-ResNet \cite{narazani2022} & P & 0.951 & 0.890 \\
3D-ViT \cite{singla2022} & M & 0.936 & 0.862 \\
3D-ViT \cite{singla2022} & P & 0.935 & 0.887 \\
ResNet early fusion \cite{song2021} & M + P & 0.855 & 0.826 \\
ResNet middle fusion \cite{narazani2022} & M + P & 0.881 & 0.826 \\
ResNet late fusion \cite{narazani2022} & M + P & 0.967 & 0.897 \\
MMTFN \cite{miao2024} & M + P & 0.936 & 0.887 \\
DiaMond \cite{li2025} & M + P & \textbf{0.971} & \textbf{0.924} \\
\textbf{Ours} & M + P & 0.962 & 0.909 \\
\bottomrule
\end{tabular}
\end{table}

Our model sits within the leading cluster (Table~\ref{tab:sota}). DiaMond \cite{li2025} and ResNet late fusion \cite{narazani2022} report marginally higher AUC; our balanced accuracy exceeds all listed methods except DiaMond. Given a test set of 110 subjects, these differences fall within overlapping confidence intervals and we do not claim superiority. The relevant point is that a 2.5D architecture with a frozen-trunk VGG backbone reaches parity with 3D CNNs and multimodal transformers at a fraction of the computational cost --- which makes the exhaustive perturbation analyses in Sections~\ref{sec:shortcut}--\ref{sec:attrib} tractable. Occlusion sensitivity over six streams and multiple ROI configurations would be prohibitively expensive on a 3D transformer.

\subsection{Error Structure}

\begin{table}[!t]
\renewcommand{\arraystretch}{1.2}
\caption{Confusion Matrix, Fusion Model, Test Set}
\label{tab:cm}
\centering
\begin{tabular}{lcc}
\toprule
 & Actual AD & Actual CN \\
\midrule
Predicted AD & TP $= 41$ & FP $= 5$ \\
Predicted CN & FN $= 5$  & TN $= 59$ \\
\bottomrule
\end{tabular}
\end{table}

Sensitivity is 89.1\% (41/46), specificity 92.2\% (59/64), and the error distribution is symmetric (Table~\ref{tab:cm}). Symmetry matters because a model that has latched onto a class-correlated confound typically produces asymmetric errors. It also indicates that the decision threshold is well placed without post-hoc adjustment, despite the 324:230 class imbalance in the cohort.

\subsection{Modality Ablation}

\begin{table}[!t]
\renewcommand{\arraystretch}{1.2}
\caption{Modality Ablation on the Fusion Model}
\label{tab:modality}
\centering
\begin{tabular}{lc}
\toprule
Mode & AUC \\
\midrule
Without MRI & 0.932 \\
Without PET & 0.938 \\
All & \textbf{0.962} \\
\bottomrule
\end{tabular}
\end{table}

Suppressing either modality at inference degrades the fusion model, and the degradation is comparable in magnitude for both (0.030 and 0.024; Table~\ref{tab:modality}). Neither modality is decorative. That the fusion model with MRI suppressed (0.932) performs slightly below the dedicated PET-only model (0.936), and similarly for the mirror case, indicates the fusion head has learned cross-modal interaction terms that are not recoverable when one modality is zeroed.

\subsection{Preprocessing Ablation}

\begin{table}[!t]
\renewcommand{\arraystretch}{1.2}
\caption{Preprocessing Ablation}
\label{tab:preproc}
\centering
\begin{tabular}{lc}
\toprule
Mode & AUC \\
\midrule
MRI - with preprocessing    & 0.944 \\
MRI - without preprocessing & 0.860 \\
PET - with preprocessing    & 0.936 \\
PET - without preprocessing & 0.930 \\
Fusion - with preprocessing    & 0.962 \\
Fusion - without preprocessing & 0.934 \\
\bottomrule
\end{tabular}
\end{table}

The asymmetry in Table~\ref{tab:preproc} is the most interpretable result in the ablation set. Removing preprocessing costs MRI 0.084 AUC but costs PET only 0.006. MRI depends critically on bias correction, skull stripping, and spatial normalization, because its discriminative signal is \emph{morphological}: a structure must be in a consistent location and free of intensity inhomogeneity for its size and shape to be comparable across subjects. PET's discriminative signal is \emph{intensity-based}, regional hypometabolism relative to a reference, and survives substantial spatial misalignment because the deficits are large and diffuse.

This also implies that an MRI pipeline without proper normalization will appear to work (0.860 is not a failure) while delivering a substantially degraded and less anatomically grounded model. The fusion model loses 0.028 without preprocessing, tracking its MRI component.

\subsection{Shortcut Controls and the Permutation Null}
\label{sec:shortcut}

\begin{table}[!t]
\renewcommand{\arraystretch}{1.2}
\caption{Brain-Removal Shortcut Controls}
\label{tab:shortcut}
\centering
\begin{tabular}{lc}
\toprule
Condition & AUC \\
\midrule
Intact     & 0.962 \\
Silhouette & 0.622 \\
Exterior   & 0.608 \\
Blank      & 0.500 \\
\bottomrule
\end{tabular}
\end{table}

\begin{table}[!t]
\renewcommand{\arraystretch}{1.2}
\caption{Label Permutation Null}
\label{tab:perm}
\centering
\begin{tabular}{lc}
\toprule
Condition & AUC \\
\midrule
Intact & 0.962 \\
Label-permutation null & 0.456 \\
\bottomrule
\end{tabular}
\end{table}

The blank condition returns exactly 0.500 (Table~\ref{tab:shortcut}), confirming the evaluation harness has no leakage path independent of image content. Silhouette (0.622) and exterior (0.608) retain some discriminative power but lose approximately 76\% and 77\% respectively of the model's above-chance performance. This residual is expected: global brain volume and ventricular expansion genuinely differ between AD and CN, and a silhouette carries coarse volumetric information. It is nonetheless small enough to exclude the hypothesis that the model is primarily reading head shape or acquisition geometry.

The PET exterior residual is the one result we flag as incompletely explained. Because PET is not skull-stripped, the exterior condition retains scalp and skull-adjacent tracer uptake as well as the outer intensity envelope, and some of the 0.608 is likely attributable to global count-rate and reconstruction differences that correlate with scanner generation and therefore, indirectly, with cohort composition. We report it rather than suppress it.

The label-permutation null (Table~\ref{tab:perm}) is the strongest of these controls. Trained under identical conditions on shuffled labels, the model reaches AUC 0.456, statistically indistinguishable from chance and, if anything, slightly below it, as expected when a network overfits noise in the training set. The architecture and training procedure generate no discrimination in the absence of real class structure.

\subsection{Forward ROI Ablation: Necessity}

\begin{table}[!t]
\renewcommand{\arraystretch}{1.2}
\caption{Forward Ablation, MRI Streams}
\label{tab:fwd_mri}
\centering
\small
\begin{tabular}{lcccc}
\toprule
ROI masked out & Area (px) & $\Delta$AUC & $\Delta$logit$_\mathrm{AD}$ & $Z$ vs.\ null \\
\midrule
Medial temporal   & 4{,}361 & 0.038 & 0.283 & $+20.3$ \\
Hippocampus       & 1{,}844 & 0.015 & 0.087 & $+5.4$ \\
Amygdala          & 1{,}011 & 0.012 & 0.057 & $+3.0$ \\
Parahippocampal   & 1{,}551 & 0.006 & 0.078 & $+4.7$ \\
Control composite & 4{,}297 & 0.005 & 0.000 & $-1.3$ \\
Precentral gyrus  & 3{,}218 & 0.002 & 0.002 & $-1.2$ \\
\bottomrule
\end{tabular}
\end{table}

\begin{table}[!t]
\renewcommand{\arraystretch}{1.2}
\caption{Forward Ablation, PET Streams}
\label{tab:fwd_pet}
\centering
\small
\begin{tabular}{lcccc}
\toprule
ROI masked out & Area (px) & $\Delta$AUC & $\Delta$logit$_\mathrm{AD}$ & $Z$ vs.\ null \\
\midrule
Posterior DMN       & 6{,}672 & 0.060  & 0.446 & $+25.9$ \\
Precuneus           & 1{,}878 & 0.016  & 0.199 & $+11.1$ \\
Posterior cingulate & 1{,}728 & 0.011  & 0.274 & $+15.6$ \\
Inferior parietal   & 3{,}066 & 0.005  & 0.126 & $+6.8$ \\
Control composite   & 3{,}088 & 0.005  & 0.036 & $+1.4$ \\
Precentral gyrus    & 1{,}801 & $-0.001$ & 0.023 & $+0.7$ \\
\bottomrule
\end{tabular}
\end{table}

The AUC changes in Tables~\ref{tab:fwd_mri} and \ref{tab:fwd_pet} are small in absolute terms, exactly as redundancy predicts. The \emph{$Z$ statistic against the spatial null is the informative column}, because it accounts for ROI area, the dominant nuisance variable, since masking more pixels perturbs the model more regardless of which pixels they are.

On that measure the separation is unambiguous. In MRI, the medial temporal composite reaches $Z = +20.3$ and every individual medial temporal structure is significant (hippocampus $+5.4$, amygdala $+3.0$, parahippocampal $+4.7$), while the area-matched control composite (4{,}297\,px, nearly identical to the medial temporal composite's 4{,}361\,px) sits at $Z = -1.3$ and the precentral gyrus at $-1.2$. Both controls are \emph{below} the spatial null, meaning masking them perturbs the model less than masking a random brain region of the same size.

In PET, the posterior DMN composite reaches $Z = +25.9$, with precuneus $+11.1$, posterior cingulate $+15.6$, and inferior parietal $+6.8$, against a control composite at $+1.4$ and precentral gyrus at $+0.7$.

The $\Delta$logit$_\mathrm{AD}$ column adds a second layer of evidence. Masking a genuinely AD-informative region should move the model's AD evidence specifically, not merely add noise. For the MRI control composite, $\Delta$logit$_\mathrm{AD}$ is 0.000 despite a $\Delta$AUC of 0.005, the region perturbs the output without moving the AD decision variable at all. Compare medial temporal (0.283) and posterior DMN (0.446). For PET, posterior cingulate is instructive: its $\Delta$AUC (0.011) is below that of precuneus (0.016), but its $\Delta$logit$_\mathrm{AD}$ (0.274) is substantially higher, indicating a small region with concentrated influence on the AD decision variable. The two metrics measure different things, and reporting only the former would have obscured this.

\subsubsection{The Double Dissociation}
Medial temporal effects are large in MRI and the corresponding PET effects are not the drivers; posterior DMN effects are large in PET and are not carried by MRI. This is precisely the pattern AD neuroimaging predicts: structural MRI indexes medial temporal neurodegeneration \cite{frisoni2010}, FDG-PET indexes posterior cingulate and precuneus hypometabolism \cite{nordberg2010}, and the two are complementary rather than redundant \cite{dukart2011}. The model has arrived at the modality-specific division of labor that the biology specifies, without being told to.

\subsection{Reverse ROI Ablation: Sufficiency}

\begin{table}[!t]
\renewcommand{\arraystretch}{1.2}
\caption{Reverse Ablation, MRI Streams (ROI Retained, All Else Masked)}
\label{tab:rev_mri}
\centering
\begin{tabular}{lcccc}
\toprule
ROI retained & Area (px) & AUC & Retained & $\rho$ \\
\midrule
Medial temporal      & 4{,}361 & 0.912 & 89.2\%   & $+0.80$ \\
Amygdala             & 1{,}011 & 0.877 & 81.8\%   & $+0.73$ \\
Hippocampus          & 1{,}844 & 0.852 & 76.2\%   & $+0.73$ \\
Parahippocampal      & 1{,}551 & 0.789 & 62.7\%   & $+0.57$ \\
Control composite    & 4{,}297 & 0.476 & $-5.2$\% & $+0.01$ \\
Control composite XL & 8{,}637 & 0.646 & 31.6\%   & $+0.26$ \\
\bottomrule
\end{tabular}
\end{table}

\begin{table}[!t]
\renewcommand{\arraystretch}{1.2}
\caption{Reverse Ablation, PET Streams (ROI Retained, All Else Masked)}
\label{tab:rev_pet}
\centering
\begin{tabular}{lcccc}
\toprule
ROI retained & Area (px) & AUC & Retained & $\rho$ \\
\midrule
Posterior DMN        & 6{,}672 & 0.865 & 79.0\% & $+0.77$ \\
Precuneus            & 1{,}878 & 0.833 & 72.0\% & $+0.66$ \\
Posterior cingulate  & 1{,}728 & 0.816 & 68.4\% & $+0.66$ \\
Inferior parietal    & 3{,}066 & 0.784 & 61.5\% & $+0.67$ \\
Control composite    & 3{,}088 & 0.593 & 20.1\% & $+0.12$ \\
Control composite XL & 4{,}716 & 0.561 & 13.2\% & $+0.09$ \\
\bottomrule
\end{tabular}
\end{table}

``Retained'' expresses the fraction of the intact model's above-chance discrimination, $(\mathrm{AUC} - 0.5)/(0.962 - 0.5)$, that survives when only the named region is visible. $\rho$ is the correlation between the reverse-ablated model's predicted probabilities and those of the intact model, a measure of whether the region reproduces the model's \emph{decisions}, not merely its accuracy.

These results (Tables~\ref{tab:rev_mri} and \ref{tab:rev_pet}) are considerably stronger than the forward ablation, in the expected direction: sufficiency tests are immune to the redundancy that suppresses necessity tests.

In MRI, the medial temporal composite alone retains 89.2\% of discriminative power (AUC 0.912 from 4{,}361 visible pixels out of 50{,}176) with $\rho = +0.80$. Every constituent structure retains more than 62\%. The area-matched control composite, with 4{,}297 visible pixels, produces AUC 0.476, \emph{below chance}, with $\rho = +0.01$, meaning its predictions are uncorrelated with the intact model's. Doubling the control area to 8{,}637 pixels raises it only to 0.646 (31.6\%, $\rho = +0.26$), less than half of what the smaller medial temporal composite achieves.

In PET the pattern repeats: posterior DMN retains 79.0\% with $\rho = +0.77$, while the area-matched control retains 20.1\% with $\rho = +0.12$, and the enlarged control performs \emph{worse} (13.2\%) despite 53\% more visible area. Adding non-informative cortex does not help; the information is regionally specific.

The amygdala result in MRI deserves comment: 1{,}011 pixels retain 81.8\% of discriminative power, a higher ratio than the larger hippocampus ROI (1{,}844\,px, 76.2\%). Amygdalar atrophy in AD is well established and often underweighted relative to hippocampal measures; the model appears to find it at least as informative per unit area.

\begin{figure*}[!t]
\centering
\includegraphics[width=\textwidth]{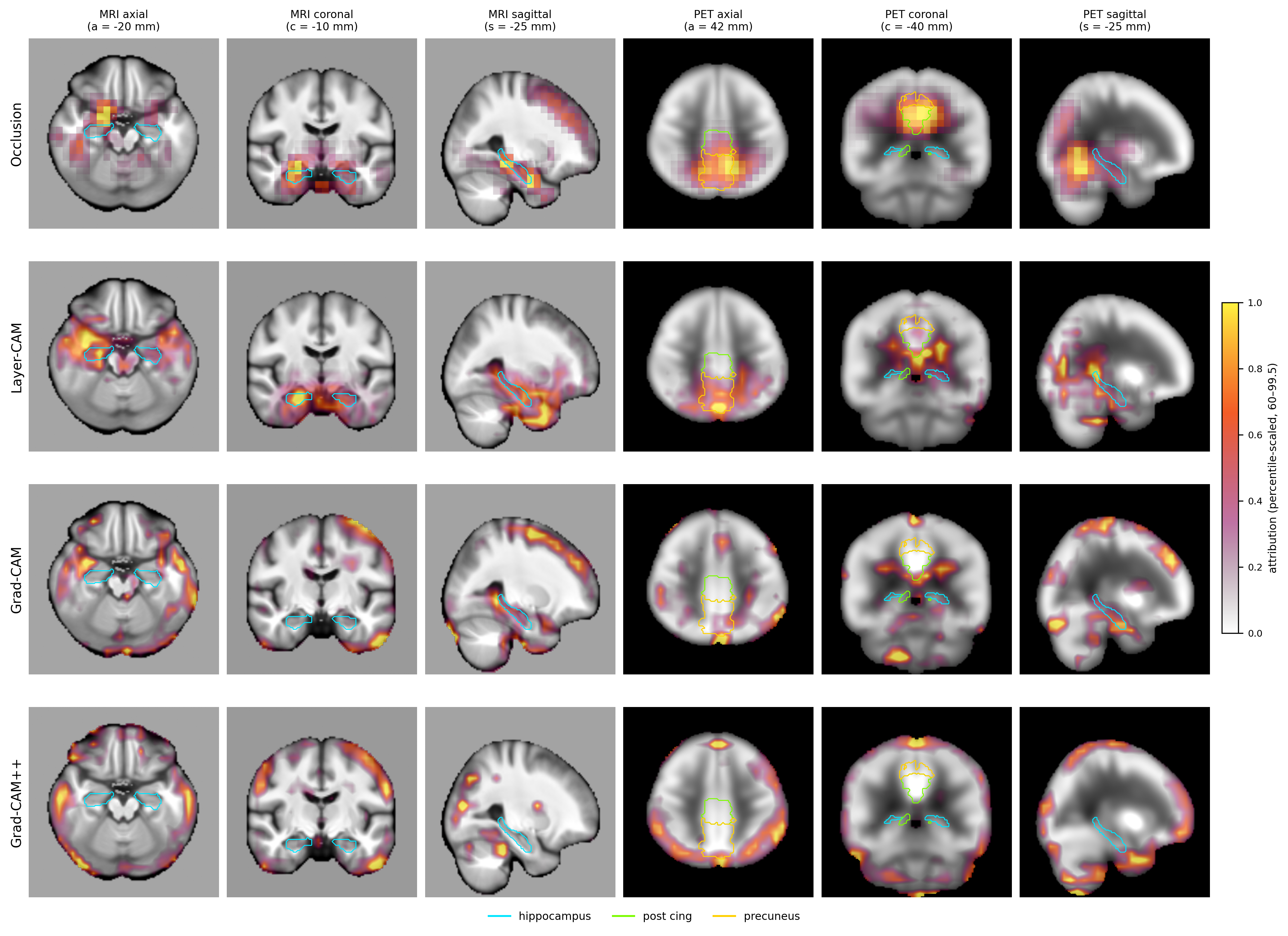}
\caption{Attribution maps for one AD subject across all six streams and four
methods, overlaid on the MNI template, with hippocampus (cyan), posterior
cingulate (green) and precuneus (orange) outlined. Occlusion and
Layer-CAM concentrate inside the outlined regions in the
corresponding modality; Grad-CAM and Grad-CAM++
concentrate on the cortical rim and image border.}
\label{fig:attr}
\end{figure*}

\subsection{Attribution Method Comparison}
\label{sec:attrib}

Fig.~\ref{fig:attr} shows attribution maps for a representative AD subject across all six streams and all four methods, with hippocampus, posterior cingulate, and precuneus outlined. The qualitative impression, occlusion and Layer-CAM concentrating inside the outlines, Grad-CAM and Grad-CAM++ concentrating along cortical rims and image edges, is confirmed quantitatively below.

\begin{table*}[!t]
\renewcommand{\arraystretch}{1.2}
\caption{Occlusion Enrichment by Stream (AD Subjects; $1.0 =$ Chance)}
\label{tab:enrich}
\centering
\begin{tabular}{lccc}
\toprule
Stream & Medial temporal (hipp / amyg / parahipp) & Posterior DMN (post-cing / precuneus / inf-par) & Control (precentral / occipital / lingual) \\
\midrule
MRI axial    & 1.43 / \textbf{2.94} / 1.17 & n/a & --- / 0.10 / 0.35 \\
MRI coronal  & \textbf{3.54 / 4.17 / 3.54} & n/a & 0.34 / --- / --- \\
MRI sagittal & \textbf{3.22 / 2.99 / 2.65} & n/a & 0.78 / 0.55 / 1.69 \\
PET axial    & n/a & 2.46 / \textbf{3.83} / 0.70 & 0.43 / 0.29 / --- \\
PET coronal  & 0.38 / --- / 0.61 & \textbf{4.99 / 3.96} / 1.15 & --- / --- / 0.18 \\
PET sagittal & 1.96 / 1.20 / 1.51 & n/a & 0.11 / 1.08 / 3.26 \\
\bottomrule
\end{tabular}
\end{table*}

\begin{table*}[!t]
\renewcommand{\arraystretch}{1.2}
\caption{Attribution Method Comparison}
\label{tab:methods}
\centering
\begin{tabular}{lcccl}
\toprule
Method & Peak enrichment, AD regions & Enrichment, control regions & Pointing game (AD, best stream) & Usable \\
\midrule
Occlusion   & $3.5 - 5.0$        & $0.10 - 0.43$ & 54.3\% (chance 16.7\%) & Yes - primary evidence \\
Layer-CAM   & $2.0 - 3.0$        & $0.22 - 0.62$ & 69.6\% (chance 6.2\%)       & Directionally; weak AD--CN separation \\
Grad-CAM    & $\approx 1.0 - 2.3$ & $0.36 - 2.00$ & 21.7\% (chance 4.8\%)       & No \\
Grad-CAM++  & $\approx 1.0$      & $0.59 - 1.32$ & 8.7\% (chance 16.7\%)       & No \\
\bottomrule
\end{tabular}
\end{table*}

\begin{table}[!t]
\renewcommand{\arraystretch}{1.2}
\caption{Pointing Game: Peak Attribution Inside the A Priori AD ROI Set}
\label{tab:pointing}
\centering
\small
\begin{tabular}{llccc}
\toprule
Method & Stream & AD & CN & Chance \\
\midrule
Occlusion  & MRI axial    & 39.1\% & 17.2\% & 6.7\% \\
Occlusion  & MRI coronal  & 52.2\% & 29.7\% & 6.2\% \\
Occlusion  & MRI sagittal & 54.3\% & 46.9\% & 4.7\% \\
Occlusion  & PET axial    & 54.3\% & 40.6\% & 16.7\% \\
Occlusion  & PET coronal  & 54.3\% & 23.4\% & 13.1\% \\
Layer-CAM  & MRI coronal  & 69.6\% & 59.4\% & 6.2\% \\
Layer-CAM  & PET axial    & 69.6\% & 70.3\% & 16.7\% \\
Grad-CAM   & MRI coronal  & 0.0\%  & 0.0\%  & 6.2\% \\
Grad-CAM++ & MRI coronal  & 2.2\%  & 1.6\%  & 6.2\% \\
\bottomrule
\end{tabular}
\end{table}

The double dissociation appears again in Table~\ref{tab:enrich}, now in attribution space and independently of the ablation experiments. MRI coronal and sagittal streams show $2.65$-$4.17\times$ enrichment in medial temporal structures. PET coronal shows $4.99\times$ in posterior cingulate and $3.96\times$ in precuneus while showing \emph{below-chance} enrichment in medial temporal structures (0.38 and 0.61), the PET streams are actively not attending to the hippocampus. Control regions are almost uniformly below chance, typically $0.10$--$0.43$.

Two control values exceed 1.0: lingual gyrus in MRI sagittal (1.69) and in PET sagittal (3.26). Both occur in sagittal streams at $s = -25$\,mm, where the occipital and lingual regions sit near the slice boundary and the occlusion window necessarily overlaps adjacent structures. We note this as a limitation of slice-plane ROI projection rather than as evidence of lingual involvement.

The ranking in Tables~\ref{tab:methods} and \ref{tab:pointing} is unambiguous and, for the most widely used method in medical imaging XAI, unflattering.

\emph{Occlusion} is the only method that satisfies both criteria. It achieves high enrichment in AD regions ($3.5$--$5.0\times$) with strongly suppressed control-region enrichment ($0.10$--$0.43\times$), and it separates AD from CN in the pointing game across every stream, most sharply in MRI coronal (52.2\% vs.\ 29.7\%) and PET coronal (54.3\% vs.\ 23.4\%), both against chance rates near 6\% and 13\%. That AD-CN gap is what makes the map an explanation of the \emph{decision} rather than a generic anatomical prior.

\emph{Layer-CAM} achieves the highest raw pointing-game rate (69.6\%) but fails the discrimination criterion: in PET axial it scores 69.6\% for AD and 70.3\% for CN, i.e.\ it points at the posterior DMN regardless of class. Its control-region enrichment ($0.22$--$0.62$) is also higher than occlusion's. It is useful as a corroborating visualization but cannot stand as evidence on its own.

\emph{Grad-CAM} fails outright. Its control-region enrichment reaches 2.00, higher than its AD-region enrichment in some streams, and in MRI coronal its pointing-game rate is \textbf{0.0\% for both classes}, i.e.\ the peak attribution never once landed inside the a priori AD ROI set across 110 test subjects. \emph{Grad-CAM++} is worse: 8.7\% against a chance rate of 16.7\%, meaning it points at AD regions \emph{less often than random}.

The mechanism is visible in \texttt{fig4.png}. Both Grad-CAM variants concentrate attribution along the cortical rim and the brain-background boundary. With a VGG16-BN trunk at $224 \times 224$, the final convolutional feature map is $7 \times 7$; each cell covers roughly $32 \times 32$ input pixels, comparable to the entire hippocampal cross-section in these slices. Gradient-weighted upsampling from that resolution cannot localize a structure of that size, and the resulting maps are dominated by high-gradient intensity boundaries. Layer-CAM's use of earlier layers \cite{jiang2021} is exactly the fix for this failure mode, and its improved pointing-game performance confirms the diagnosis, though it does not fix the class-discrimination problem.

The practical implication is uncomfortable for a large body of published work: a Grad-CAM figure showing plausible-looking highlights over medial temporal cortex constitutes no evidence that a model is using medial temporal information. In our model, which the ablation experiments prove \emph{does} use medial temporal information, Grad-CAM's peak lands inside those regions 0\% of the time. A method that fails to detect a genuine dependency cannot be used to confirm one.

\section{Discussion}

\subsection{What the Evidence Ladder Establishes}

Taken together the experiments support a chain of claims, each with its own control. The model is not exploiting non-brain content: silhouette, exterior, and blank conditions lose 76\%, 77\%, and 100\% of above-chance performance respectively. The model is not exploiting optimization artifacts: the label-permutation null returns AUC 0.456. Both modalities contribute: suppressing either costs 0.024-0.030 AUC, and their contributions are not interchangeable. AD-relevant regions are causally necessary: masking them shifts the AD logit far beyond the spatial null ($Z$ up to $+25.9$) while area-matched controls do not ($Z = -1.3$ to $+1.4$). AD-relevant regions are largely sufficient: the medial temporal lobe alone retains 89.2\% of discrimination in MRI and the posterior DMN alone retains 79.0\% in PET, versus $\leq 31.6$\% for area-matched and enlarged controls. And the dependency is modality-specific in the direction AD biology predicts, confirmed independently by ablation and by occlusion attribution.

No single one of these is decisive. Their conjunction is considerably harder to explain by any account other than that the model is reading AD neuropathology.

\subsection{Redundancy and the Limits of Necessity Testing}

A methodological finding with implications beyond this study: in a six-stream architecture, forward ablation of individual regions produces near-zero $\Delta$AUC, not because those regions are unimportant but because the remaining streams compensate. Had we reported only forward ablation, we would have concluded that no region matters, the exact opposite of the truth, which reverse ablation makes clear.

Any redundant architecture, multi-view, multi-modal, or ensemble, will exhibit this. We suggest that necessity-only ablation studies on such architectures be interpreted as lower bounds, and that sufficiency testing with area-matched controls is the more informative design. The $\rho$ column in Tables~\ref{tab:rev_mri} and \ref{tab:rev_pet} strengthens this further by asking whether a region reproduces the model's \emph{decisions} and not just its accuracy: the control composite's $\rho = +0.01$ in MRI shows that its 0.476 AUC is not merely poor but entirely unrelated to what the intact model does.

\subsection{The Status of Gradient-Based Attribution}

Our results place Grad-CAM and Grad-CAM++ in a category that the XAI methodology literature has warned about \cite{vandervelden2022,holzinger2019} but that AD neuroimaging practice has largely ignored: methods that produce confident, plausible, publishable visualizations while failing every quantitative test of faithfulness. The 0.0\% pointing-game rate for Grad-CAM in MRI coronal is the clearest single data point in this paper.

This is not a claim that Grad-CAM is broken in general, it was designed for and validated on natural-image classification with large, centrally located objects \cite{selvaraju2020}. It is a claim that its assumptions fail for medical imaging with small subcortical targets and final feature maps at $7 \times 7$ resolution. Layer-CAM's hierarchical formulation \cite{jiang2021} partially addresses this. Perturbation-based methods \cite{zeiler2014}, despite their computational cost, remain the ones whose output has a causal interpretation by construction, and we recommend that attribution claims in this domain be anchored to perturbation evidence with quantitative enrichment statistics and explicit class-discrimination checks, not to heatmap figures.

\subsection{Clinical Framing}

A classifier that separates established AD from CN is not itself clinically useful, that distinction is already made confidently at the bedside. The value of this class of model lies in what it enables downstream: MCI conversion prediction, differential diagnosis among dementias, and quantitative staging. Each requires far more trust than an AUC provides, and trust of that kind requires evidence of the sort assembled here. The causability framing of Holzinger \emph{et al.} \cite{holzinger2019} is the relevant standard: the question is not whether the system produces an explanation, but whether a clinician can verify that the explanation is grounded in the same evidence they would use. A demonstration that the model's hippocampal dependency is carried by MRI and its posterior cingulate dependency by PET is legible in exactly that way.

\section{Limitations}

\textbf{2.5D representation.} Six fixed slices, however well chosen, discard most of each volume. This constrains the ceiling on performance and means regional analyses are confined to the slice planes used. It is also what makes the perturbation analyses computationally feasible; the trade-off is explicit.

\textbf{Single cohort.} All data come from ADNI \cite{petersen2010}. External validation on an independent cohort with different scanners and demographics is required before generalization claims, and site-invariance approaches such as those of Belhaj Ali \emph{et al.} \cite{belhajali2025} are the natural direction.

\section{Conclusion}

We presented a six-stream 2.5D MRI-PET fusion network for AD versus CN classification that reaches AUC 0.962, accuracy 0.909, and F1 0.891 on a strictly subject-disjoint, physically separated ADNI test set, with performance competitive against 3D CNN and multimodal transformer baselines at substantially lower computational cost.

The principal contribution is the evidence assembled around that result. Shortcut controls and a label-permutation null exclude non-brain and artifactual explanations. Forward ROI ablation demonstrates that AD-relevant regions are causally necessary against a spatial null that area-matched controls do not exceed. Reverse ROI ablation demonstrates that they are largely sufficient, with the medial temporal lobe alone retaining 89.2\% of MRI discrimination and the posterior default-mode network alone retaining 79.0\% of PET discrimination. A quantitative comparison of four attribution methods shows that occlusion sensitivity is the only one that both concentrates inside AD regions and separates AD from CN, while Grad-CAM and Grad-CAM++ fail entirely, in one stream, Grad-CAM's peak attribution never lands inside the a priori AD region set across the whole test cohort.

Across ablation and attribution alike, the same double dissociation emerges: hippocampal and medial temporal evidence carried by MRI, posterior cingulate and precuneus evidence carried by PET. That the model recovers the modality-specific division of labor established by decades of AD neuroimaging, without supervision toward it, is stronger evidence of biological validity than any accuracy figure.

We suggest that the components of this protocol, subject-level splitting with a physically separated test set, shortcut controls, permutation nulls, area-matched forward and reverse regional ablation, and quantitative attribution benchmarking with class-discrimination checks, be treated as reporting requirements rather than optional extras.

\section*{Acknowledgment}

Data collection and sharing for this project was funded by the Alzheimer's Disease Neuroimaging Initiative (ADNI) (National Institutes of Health Grant U01 AG024904). ADNI investigators contributed to the design and implementation of ADNI and provided data but did not participate in the analysis or writing of this report.

\end{document}